\documentclass[conference]{IEEEtran}
\IEEEoverridecommandlockouts
\usepackage{cite}
\usepackage{amsmath,amssymb,amsfonts}
\usepackage{algorithmic}
\usepackage{graphicx}
\usepackage{textcomp}
\usepackage{xcolor}
\usepackage{subfig}

\usepackage{threeparttable}
\usepackage{booktabs}

\renewcommand{\paragraph}[1]{\textbf{#1} }

\def\BibTeX{{\rm B\kern-.05em{\sc i\kern-.025em b}\kern-.08em
    T\kern-.1667em\lower.7ex\hbox{E}\kern-.125emX}}
\begin{document}

\title{Towards Fully Intelligent Transportation through Infrastructure-Vehicle Cooperative Autonomous Driving: Challenges and Opportunities}

\author{\IEEEauthorblockN{
Shaoshan Liu\IEEEauthorrefmark{1},
Bo Yu\IEEEauthorrefmark{1},
Jie Tang\IEEEauthorrefmark{2}\IEEEauthorrefmark{4}, 
Qi Zhu\IEEEauthorrefmark{3}\\}
\IEEEauthorblockA{\IEEEauthorrefmark{1}PerceptIn, U.S.A.\\}
\IEEEauthorblockA{\IEEEauthorrefmark{2}South China University of Technology, China\\}
\IEEEauthorblockA{\IEEEauthorrefmark{3}Northwestern University, U.S.A.\\}
\IEEEauthorblockA{\IEEEauthorrefmark{4}corresponding author cstangjie@scut.edu.cn}}

\maketitle

\begin{abstract}
The infrastructure-vehicle cooperative autonomous driving approach depends on the cooperation between intelligent roads and intelligent vehicles. This approach is not only safer but also more economical compared to the traditional on-vehicle-only autonomous driving approach. In this paper, we introduce our real-world deployment experiences of cooperative autonomous driving, and delve into the details of new challenges and opportunities. Specifically, based on our progress towards commercial deployment, we follow a three-stage development roadmap of the cooperative autonomous driving approach:infrastructure-augmented autonomous driving (IAAD), infrastructure-guided autonomous driving (IGAD), and infrastructure-planned autonomous driving (IPAD).
\end{abstract}

\maketitle

\section{Introduction}
\label{sec:intro}
In the past few years, we have developed technologies with both the on-vehicle only autonomous driving approach as well as the infrastructure-vehicle cooperative autonomous driving approach. The cooperative autonomous driving approach depends on the cooperation between intelligent roads and intelligent vehicles. Specifically, our commercial cooperative autonomous driving system consists of roadside-deployed intelligent systems, dubbed Systems-on-Road (SoRs), and on-vehicle autonomous driving systems, dubbed System-on-Vehicle (SoV), such as the one presented in \cite{yu2020building}. Through our real-world deployment experiences, we find that the cooperative driving approach is safer, more efficient, and more economical compared to the traditional on-vehicle-only autonomous driving approach. 

\subsection{Safety}

According to the U.S. Centers of Disease Control and Prevention \cite{centers2019road}, 1.35 million people die in road crashes each year, and an additional 20 million suffer non-fatal injuries, often resulting in long-term disabilities. Regarding why traffic accidents happen, human drivers have a crash rate of 4.2 accidents per million miles (PMM), and the current autonomous vehicle crash rate is 3.2 crashes PMM \cite{blanco2016automated}. While autonomous vehicles deliver a significant safety improvement compared to the human counterpart, there are corner cases, such as blind spots, that neither human drivers nor autonomous vehicles can handle without the help of intelligent infrastructures. On the other hand, the cooperation between SoVs and SoRs provides a comprehensive global view of the traffic condition, thus not only eliminating blind spots but also greatly extending SoVs perception distance to provide more reaction time for autonomous vehicles. At our deployment site, with blind-spot elimination and extended perception from SoRs, we have observed that \textbf{the autonomous vehicle disengagement rate has been dropped by more than 90\%}. Eventually, we aim to drop the vehicle crash rate to zero through cooperative autonomous driving.

\subsection{Efficiency}

Traffic inefficiency, such as congestion, imposes very high costs on our society.  For instance, for trucking industry alone, the American Transportation Research Institute estimates that congestion costs the U.S. \$74.1 billion annually, of which \$66.1 billion occurs in urban areas \cite{hooper2018cost}. In addition, there are other costs such as pollution and accidents, all of which cost each city billions of dollars every year. Using intelligent SoRs, we can obtain a comprehensive and fine-grained information of the traffic condition. For instance, at our deployment site, utilizing per-vehicle tracking data generated by our deployed SoRs, we are able to \textbf{improve traffic efficiency by 40\%} through intelligent traffic light control. 

\subsection{Cost}

There is an estimated 1.2 billion vehicles on world's roads today and this is expected to grow to 2 billion by 2035.  On the other hand, in the U.S. the road capacity is designed to process 1,900 vehicles per hour per lane, and there are only about 4 million miles of public road in the U.S. \cite{national2005assessing}. If we were to convert all cars in the world into autonomous vehicles, it would incur an enormous social cost, as autonomous vehicles are still very expensive to build \cite{liu2020autonomous}. A better solution is to invest in the road infrastructure and making it intelligent. Gradually, as SoRs become more powerful, we can migrate more workloads from the SoVs to the SoRs, hence greatly reducing the hardware and energy costs of autonomous vehicle deployments. Eventually, we aim to have autonomous vehicles equipped with only basic perception and control capabilities, relying on the SoRs for proactive perception and planning tasks.  We estimate that this will \textbf{drop the cost of autonomous vehicles by more than 50\%}.

\section{Infrastructure-Vehicle Cooperative Autonomous Driving System Overview}
\label{sec:overview}

\begin{figure}[t]
\centering
\includegraphics[width=1\columnwidth]{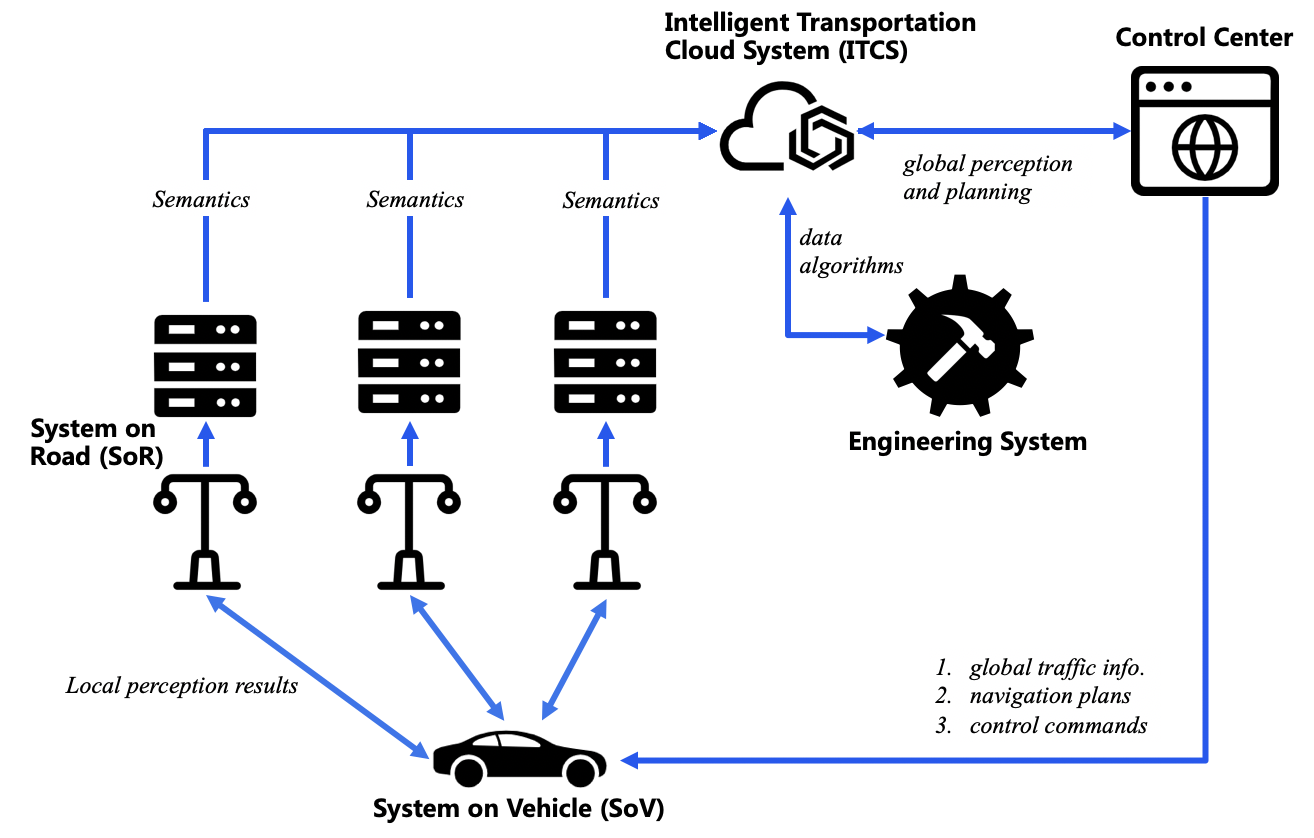}
\caption{system architecture of cooperative driving}
\label{fig:arch}
\end{figure}

Figure \ref{fig:arch} presents an overview of our cooperative autonomous driving system. It consists of the SoVs, the SoRs, the intelligent transportation cloud system (ITCS), the engineering system, and the control center: the SoRs provide local perception results to the SoVs for blind spot elimination and extended perception \textbf{to improve safety}; meanwhile the SoRs process incoming sensor data and send the extracted semantic data to the ITCS for further processing;  the ITCS fuse all incoming semantic data to generate global perception and planning information, then the control center can dispatch global traffic information, navigation plans, and control commands to the SoVs \textbf{to achieve optimal traffic efficiency}. The engineering system consumes data collected from the SoRs and the SoVs and produce periodic algorithm updates for the SoRs, the SoVs, and the ITCS.  The purpose of the engineering system is \textbf{to maximize development productivity and to minimize development and testing costs}. In the rest of this paper, we are going to review the key components in the cooperative autonomous driving system, as well as the development roadmap. 

\section{Design of the System-on-Road (SoR)}
\label{sec:sor}

Our SoVs are full-fledged level-4 autonomous vehicles, equipped with mmWave radars, cameras, LiDARs, GPS, IMU, and a on-board computer. The key workloads running on the SoVs include sensing, perception, localization, tracking, prediction, planning, and control. Over 80\% of the computing power and hardware cost are dedicated to the sensing, perception, and localization modules, making these the perfect target for offloading to the SoRs \cite{liu2020creating}.

On the other hand, the SoRs are designed to enrich the SoVs' perception and to offload sensing and computing from the SoVs. Each SoR is equipped with three cameras, three radars, two LiDARs, and an edge computer, and a RSU for C-V2X communication. To enable cooperative driving, the SoVs need to communicate with the SoRs.  Thus, each SoV is equipped with an C-V2X on-board unit (OBU), whereas each SoR is equipped with an C-V2X road-side unit (RSU), the RSUs and OBUs communicate directly with each other to facilitate the fusion of perception data from the vehicle side as well as the road side, effectively and efficiently forming a local perception map to aid the autonomous vehicles' navigation, especially to improve safety through blind spot elimination and extended distance perception.  In our current deployment, we utilize LTE-V2X for communication and will support 5G-V2X as more 5G base stations get deployed. The current communication coverage is around 300 meters with average latency less than 25 ms. 

%Although we look forward to ubiquitous 5G deployments to further reduce communication latency and improve bandwidth, the current LTE-V2X seems sufficient to sustain semantic data communication, with 5G we might be able to support raw data communication to offload perception processing from vehicle to road, and vice versa. 

The current effective SoR sensing coverage is around 250 meters (125 meters for each direction), hence we need at least 4 SoRs per kilometer to ensure full coverage. In each second, a SoR processes at least 100 MB of raw sensing data, and generates 100 KB of semantic data, with a 50 ms latency for each LiDAR and camera frame to ensure 20 updates per second. To sustain 20 updates per second output, the SoR compute system consists of an Intel i9 CPU and multiple Nvidia GTX 1080Ti GPUs, the peak power consumption of a SoR can reach 800 W. 

Besides significant road-side investments, one problem of the SoR deployments is to provide an additional 3200 W power supply per kilometer. There are different approaches to alleviate this problem: 1.) Using lightweight computing units on SoRs to reduce cost and power consumption through software optimization techniques, such as the compression-compilation co-design optimization \cite{liu2020cocopie}. Our initial results demonstrate that we might be able to use lightweight SoC, such as Nvidia Xavier, or even Qualcomm Snapdragon devices to handle road-side perception tasks. 2.) Using fiber-optic connections to link multiple SoRs to a local edge compute center to consolidate computing. Our initial results demonstrate that overall deployment cost and power consumption can be reduced significantly by consolidating and time-sharing compute resources at the local edge compute center. We are currently exploring both options, as well as the ideal combination of these approaches.

\section{Design of the Intelligent Transportation Cloud System (ITCS)}
\label{sec:cloud}

A standard autonomous driving cloud for on-vehicle-only autonomous driving provides distributed computing and storage infrastructure supports.  On top of the computing and storage infrastructure, autonomous driving cloud provides services including deep learning model training for algorithm development, distributed simulation for new algorithm verification, High-Definition (HD) map generation for localization and perception data aggregation, these are offline services because the cloud does not have to participate in real-time perception or planning tasks in on-vehicle autonomous driving \cite{liu2017unified}. 

For cooperative autonomous driving, in addition to the standard autonomous driving cloud computing tasks, ITCS requires a new online cloud processing data pipeline: first, each SoR or SoV will upload the semantic perception results to the cloud for further processing. The semantic data includes \textit{timestamp, object type, object shape, object location, object speed, object heading etc.}  Each SoR or SoV uploads about 100 KB of semantic data per second. Second, ITCS fuses all incoming data, then merges or removes duplicate entries to form a global real-time perception map, which contains all objects' dynamic information. Third, based on the global perception map, ITCS predicts each object's trajectory, monitors each lane's traffic condition, and plans optimized routes or even trajectories for each autonomous vehicle. The results are dispatched from ITCS to each autonomous vehicle to achieve maximum traffic efficiency and to guarantee traffic safety. 

One problem is how to partition compute resources on the cloud, some options include: 1.) vertical partitioning: each compute unit is responsible for all tasks (semantic data fusion, trajectory prediction, trajectory planning) for all objects within an area, assuming that the global perception map has been partitioned into different areas. 2.) horizontal partitioning: each compute unit is responsible for only one task across multiple areas.  We are currently exploring these two options with the objective of maximizing throughput while minimizing cloud resource utilization.

\section{Design of the Engineering System}
\label{sec:eng}

Simulation is the cornerstone of autonomous driving software development, in which autonomous driving functions are exercised in the virtual environment. Development and validation in simulated environment have been widely adopted by autonomous driving industry as an important complementary to physical testing to improve development efficiency and functional safety. As physical deployment, either on public roads or in closed-courses (e.g. Waymo's castle), is time consuming and expensive, building virtual scenarios of interests is commonplace to accelerate development efficiency. 

\begin{figure}[t]
\centering
\includegraphics[width=1.\columnwidth]{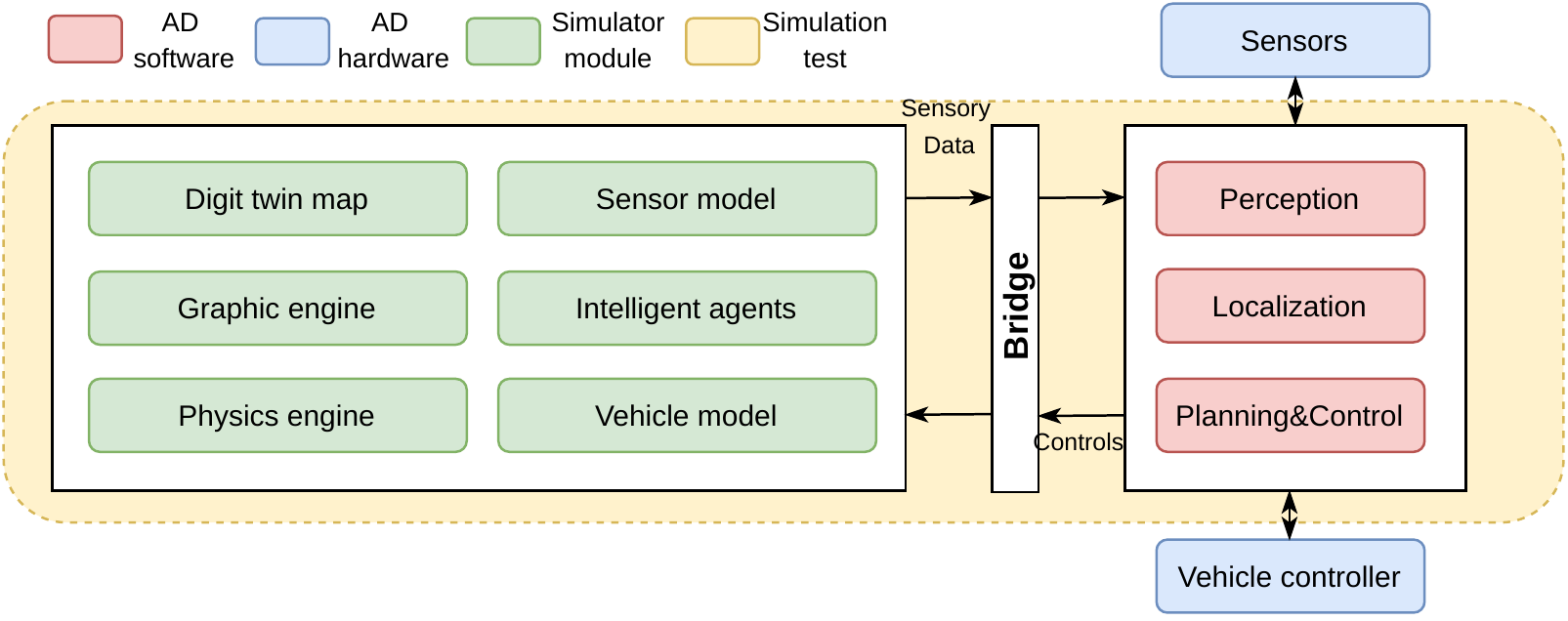}
\caption{Simulation system architecture for autonomous driving.}
\label{fig:sim}
\vspace{-5mm}
\end{figure}

\paragraph{Simulation system design}
For effective testing, the simulator needs to provide autonomous vehicles a virtual representative of the physical world, including static environment, dynamic agents (e.g. vehicles, pedestrian, cyclist), and sensing streams (e.g. camera, Lidar, Radar). To emulate the real world behaviour with high-fidelity, the following modules are fundamental building blocks of the simulation system:
\begin{itemize}
    \item Digital twin map that describes static environment and objects of interests to autonomous driving, such as lanes, traffic signs and lights. 
	\item Physical engine that simulates objects' physical behaviour and interactions with environment.
	\item Graphic engine that renders the scene to generate realistic data for emulating camera images and Lidar point clouds.  
	\item Sensor models that emulate sensing processes and data, such as camera and Lidar, based on facilities of graph and physical engine. 
	\item Vehicle model that models vehicles' behaviour given control commands and physical properties of the vehicle. 
	\item Intelligent agents that models the behaviours of other vehicles, cyclists, and pedestrians, and the interactions between agents and autonomous vehicles.
	\item Interfaces with autonomous driving system, which transfers simulated data and vehicle controls between simulator and autonomous driving system.
\end{itemize}

\paragraph{Simulation architecture}
The architecture of the autonomous driving simulation system is illustrated in Fig. \ref{fig:sim}. The simulator models the sensing mechanism, and generates sensory data by simulating interactions between sensors and environments. Exercised by the simulated sensory data, the autonomous driving software generates control commands to drive the vehicle. As a perhaps obvious note, compared with physical test, the simulation test replaces real sensor and vehicle hardware with virtual environment.

Modern game engines, such as Unity and Unreal Engine (UE), are with powerful graphic and physic engines for 3D scene rendering, collision detection, vehicle models, etc., which make them a suitable choice for building autonomous driving simulators. Many high fidelity open source (e.g. Carla\cite{Dosovitskiy17}) and commercial simulators (e.g. Nvidia Constellation) use game engines as the foundation for simulating sensors and interactions among agents. We choose UE for building our simulator as it is open-source. Based on UE, plug-ins are developed for modelling camera and Lidar sensors, 3D environment, vehicle and pedestrian's behaviour. Our simulator works in a cycle-by-cycle manner. In each cycle, the simulator updates each agent's state (such as positions, velocities, etc.) by simulating its physical interactions with the environment; the graphic engine renders the 3D environment in camera's field of view, and generates images for each camera. Ray tracing in UE is used to emulate ToF (time of flight) principal for each ray of Lidar. 

\paragraph{Cost and efficiency model} Let $n_v$, $c_p$, $s$ be the number of vehicles, the testing cost (per hour per vehicle) and the average speed ($km/h$) of a vehicle; let $n_s$, $c_s$ and $cap$ be the number of simulation server, simulation cost (per server per hour) and the capacity of the server, in terms of the number of simulation jobs. The cost and efficiency model of physical and simulation test are in Table \ref{tab:cost}. Based on real operations, the physical test costs 180 $\$/hour$; the simulation server (4 GPUs and 32 CPUs) costs 8.5 $\$/hour$, and can simultaneously run 4 simulation jobs. The cost reduction and efficiency improvement of simulation are 250x and 12x.

\begin{table}[t]
\caption{Cost and efficiency of physical vs. simulation test.}
\centering
\resizebox{0.8\columnwidth}{!}{
\renewcommand*{\tabcolsep}{3pt}
\begin{tabular}{ccc}
\toprule[0.15em]
\multicolumn{1}{l}{\textbf{}} & Physical       & Simualtion \\
\midrule[0.05em]
Cost (dollars per day per unit)        & $h_p{\times}n_{v}{\times}c_{p}$  & $h_s{\times}n_s{\times}c_s$ \\ 
Efficiency (km per day per unit)       & ${h_p}{\times}s$           & ${h_s}\times{s}\times{cap}$           \\
\bottomrule[0.15em]
\end{tabular}
}
\label{tab:cost}
\vspace{-5mm}
\end{table}

\section{Commercial Development Roadmap of Cooperative Autonomous Driving}
\label{sec:stage}

Based on our progress towards commercial deployment, we follow a three-stage commercial development roadmap of the infrastructure-vehicle cooperative autonomous driving approach as follows, with each stage focused on one key objective:

\begin{itemize}
  \item stage 1: infrastructure-augmented autonomous driving (IAAD), in which autonomous vehicles fuse both vehicle-side and infrastructure-side perception outputs to \textbf{improve the safety} of autonomous driving. In this stage, both the SoVs and the SoRs are equipped with full sensing and computing capability, and we aim to fuse the information from both the SoRs and the SoVs to improve safety and to achieve maximum computing and energy efficiency. 
  \item stage 2: infrastructure-guided autonomous driving (IGAD), in which autonomous vehicles can offload all the proactive perception tasks to the infrastructure in order to \textbf{reduce per-vehicle deployment costs}. In this stage, while the SoRs are equipped with full sensing and computing capability, we can greatly reduce the on-vehicle hardware deployment (both computing and sensing) costs. We aim to design secure, reliable, and safe infrastructure supports for autonomous vehicles.
  \item stage 3: infrastructure-planned autonomous driving (IPAD), in which the infrastructure takes care of both perception and planning, thus \textbf{achieving maximum traffic efficiency and cost efficiency}. In this stage, the SoVs are equipped with only basic sensing and computing capability, and we aim to optimize global traffic efficiency by centrally planning for the trajectory for each vehicle.
\end{itemize}

\section{Challenges and Opportunities}
\label{sec:cha}

In this section, we conclude by presenting the research challenges and opportunities in infrastructure-vehicle cooperative autonomous driving, as it evolves from IAAD, to IGAD, and eventually reaching IPAD. To facilitate cooperative autonomous driving, especially to enable SoVs to interact with the infrastructure, we have developed an on-board fusion engine, we expect to expand the functionalities of the fusion engine as we progress towards IPAD. 

\textbf{Network performance and security is the key challenge in the IAAD stage}: as an autonomous vehicle travels at a high speed, for at least ten times per second, the SoV fuses perception results from multiple SoRs along with its local perception results, and feeds the combined results to the prediction and the planning module. There is a regular deadline for receiving data from the SoRs. If a deadline is missed, the SoV has to rely on its own sensors for perception until the next deadline. Hence this process is extremely sensitive to network jitters. The fusion engine monitors the deadlines and decides whether to wait for information from the SoRs or move forward with only the SoV perception results.

In our deployments, we have observed network jitters ranging from 3 ms to 100 ms, which significantly impacts the ability of SoRs to augment the perception capability of SoVs. On the other hand, network bandwidth is not a severe constraint, as the data exchange between our SoVs and SoRs is semantic information, and the required network bandwidth is within 1.5 MB/s. Hence, cooperative driving demands very stable network connection with network jitters less than 5 ms.

More importantly, security is the most critical challenge. A hacker can attack the C-V2X network, or even the SoRs to pass incorrect information to the SoVs, potentially leading to lethal results.  Currently, we rely on traditional network security techniques to protect the C-V2X network. Besides network security techniques, the fusion engine checks whether to consume the perception results from the SoRs, in case there is a information conflict between the SoRs and the SoV, currently the fusion engine always trusts the SoV. 

\textbf{Reliability is the key challenge in the IGAD stage}: the SoVs are left with only reactive perception capability, and rely solely on the SoRs for proactive perception. The on-board fusion engine has to fuse information from multiple SoRs, in case one SoR fails, the fusion engine can still utilize data from other SoRs to guide its navigation.  In addition, in this stage we cannot solely rely on the C-V2X network, the SoV needs to also receive information 
from the regular 5G network, such that in case the C-V2X network fails, the fusion engine can switch to the regular 5G network for traffic information.  If both networks fail, the fusion engine needs to deploy a safety strategy, such as stopping the vehicle at the shoulder of the road.

\textbf{Efficiency is the key challenge in the IPAD stage}: in the IPAD stage, all semantic information from the SoRs and the SoVs are aggregated to the ITCS to form a global perception map with detailed information of each vehicle, and the ITCS generates a global navigation plan for each autonomous vehicle. The objective is to balance the traffic load on each road to achieve maximum traffic efficiency.  Then the SoRs perform local trajectory planning for each autonomous vehicle, the fusion engine is responsible for handling transitions between the SoRs, and switching to the on-board planner when the SoRs become unavailable. 

\bibliographystyle{IEEEtran}
\bibliography{ref}

\end{document}